\documentclass[times,review,12pt]{elsarticle}

\journal{Expert Systems}

\usepackage{amssymb}
\usepackage{amsmath}
\usepackage{booktabs} 
\usepackage{array}   
\usepackage{multirow}
\usepackage{array}
\usepackage{graphicx}
\usepackage{arydshln}

\usepackage{tikz}
\usetikzlibrary{positioning}

\usepackage[skip=2pt]{caption}

\begin{document}

\begin{frontmatter}


 \title{\tnoteref{label1}\tnoteref{label2}}

 \tnotetext[label1]{Corresponding author.}

\tnotetext[label2]{Under review at Expert Systems.}

\author{Daniel Vila-Cruz\corref{label1} }
\ead{d.vila1@udc.es}
\author{Laura Morán-Fernández}
\ead{laura.moranf@udc.es}
\author{Verónica Bolón-Canedo}
\ead{veronica.bolon@udc.es}

\title{Beyond Backbone Backpropagation: A Decoupled Strategy for Efficient Transfer Learning }


\affiliation{organization={CITIC, Universidade da Coruña},
            city={A Coruña},
            country={Spain}}

\begin{abstract}
Deep learning models achieve state-of-the-art image classification but face deployment challenges due to computational costs and energy demands. We propose a lightweight training strategy that adapts normalization layers of the model to the new domain and decouples feature extraction from classifier optimization, reducing overhead by precomputing features only once. A redesigned classifier head with margin-based weighted loss further minimizes ambiguity without end-to-end backpropagation. Evaluated across four CNN architectures (Res\-Net18, ResNet50, MobileNet, DenseNet121), three Transformer models (ViT, \allowbreak Swin and DeiT) and three medical datasets (Brain Cancer MRI, BreakHis and PatchCamelyon), our approach significantly reduces the required training time with only a marginal accuracy trade-off, often matching or surpassing baseline performance. This efficiency translates to reducing CO\textsubscript{2} by orders of magnitude, offering a practical and environmentally sustainable solution for resource-constrained clinical or prototyping environments.
\end{abstract}



\begin{highlights}
\item We introduce a decoupled framework eliminating backbone backpropagation.
\item We propose normalization tuning for efficient domain adaptation.
\item We develop margin-based weighted training for the decoupled head.
\item The approach achieves competitive accuracy on medical benchmarks.
\item We demonstrate training viability on CPU infrastructure.

\end{highlights}


\end{frontmatter}



\section{Introduction}


The advancements of deep learning have led to breakthroughs in various fields, achieving state-of-the-art performance. While these models yield substantial gains in accuracy, these improvements come along with higher computational complexity. For large-scale technology firms and research laboratories, the hardware requirements for such models are manageable. However, this situation contributes to a growing  digital divide, introducing disparities in technology access \cite{schwartz2020green, osonuga2025bridging, poh2025bridging}.

As a result, there exist many application domains in which the direct adoption of modern deep learning approaches is not feasible, including edge and embedded systems, mobile applications, real-time systems and robotics. In this work, we focus on a domain where hardware availability constitutes a critical constraint, specifically the deployment of novel deep learning architectures in clinical centers for image analysis. The proposed pipeline is designed around model architectures and training strategies rather than domain-specific assumptions, making it inherently domain-agnostic and transferable to other scenarios. 


There are multiple approaches to reduce these limitations, the most common one being fine-tuning an already trained model to adapt it to a new domain. However, due to the tendency of creating larger and larger models, even fine-tuning can be impractical for clinical facilities lacking specialized computational infrastructure, as they do not have the required resources. Alternative strategies focus on reducing the computational burden of models. One such method is pruning, which reduces model complexity by eliminating redundant weights \cite{liang2021pruning}. Another is quantization, which lowers the precision of weights and activations to decrease memory usage and computational overhead \cite{gholami2022survey}. A third prominent technique is knowledge distillation, in which a smaller ``student" model is trained to replicate the behavior of a larger, pre-trained ``teacher" model, achieving a model that is significant smaller than the original \cite{gou2021knowledge}. While these methods do enhance efficiency, they often come with trade-offs, such as the requirement for iterative optimization or a potential decrease in performance. 

The medical domain was specifically selected for this research, as further detailed in Section \ref{subsec:datasets}, because it exemplifies rigorous constraints for model adaption. In this context, datasets are frequently limited in size and diversity, which can lead to overfitting or the development of unwanted biases when large backbones are fully fine-tuned \cite{lepcha2025deep, zech2018variable, oakden2020hidden}.



In response to these limitations, this work introduces a novel training strategy designed for efficiency and energy consciousness. Our approach decouples feature extraction from classifier optimization, effectively eliminating the computational overhead associated with iterative backbone processing. The primary contributions of this work are as follows:

\begin{itemize}

\item \textbf{Computational reallocation}: We demonstrate that the marginal gains of full backbone backpropagation are outweighed by its significant computational overhead. By performing only minimal, lightweight adaptation of normalization layers, we generate sufficient discriminative features that a classifier with better decision boundaries can effectively partition the feature space. By reallocating the primary computational effort to the optimization of this decoupled head, our framework achieves competitive accuracies in a fraction of the time, proving that a targeted refinement of the decision-making layers can compensate for the lack of full-backbone updates.

\item \textbf{Efficient domain adaptation via normalization tuning}: We propose a mechanism to bridge the domain gap without the cost of full backbone backpropagation. By specifically targeting Batch Normalization (BN) in CNNs and Layer Normalization (LN) in Transformers, we adapt the model to new data with minimal parameter updates. 

\item \textbf{Margin-based weighted classifier training}: To improve classification performance despite a frozen backbone, we introduce a weighted training for decoupled head.  This approach assigns higher importance to ambiguous samples near the decision boundary, enhancing the model's discriminative power in data-scarce scenarios.

\item \textbf{Computational viability}: We demonstrate that our strategy significantly lowers the hardware barrier to train deep learning models. Specifically, our approach enables training on standard CPU faster than traditional methods like fine-tuning or LoRA on GPU, providing a practical alternative for facilities without dedicated high-performance computing infrastructures.

\end{itemize}



In addition to the technical challenges, there is a growing imperative for frugal and energy-aware machine learning practices \cite{schwartz2020green, strubell2020energy, verdecchia2023systematic}. These practices aim to minimize computational costs and carbon footprints without sacrificing performance. Many existing solutions frequently employ unnecessarily complex architectures to solve problems that could be addressed more efficiently with simpler models. This work directly addresses these multifaceted challenges by proposing a resource-efficient training pipeline that also allows to test different hyper parameter settings and prototype different model architectures with an important reduction of training time.

Beyond computational efficiency, our findings suggest a broader insight: in many practical transfer learning scenarios, performance degradation under domain shift appears to stem more from statistical misalignment in normalization layers than from representational inadequacy of the pretrained backbone. This observation motivates our focus on lightweight normalization adaptation rather than full parameter optimization.

The remainder of this paper is organized as follows. Section \ref{Sec:state-of-the-art} reviews the state-of-the-art of efficient training strategies. Section \ref{Sec:methodology} describes the proposed methodology. Section \ref{Sec:evaluation} explains the evaluation setup. Section \ref{Sec:results} details the experimental results. Finally, Section \ref{Sec:conclusions} summarizes the conclusions of this work.



\section{State-of-the-Art}
\label{Sec:state-of-the-art}

The rapid evolution of deep learning architectures has led to substantial improvements in performance across a wide range of computer vision tasks. However, these gains are often accompanied by increasing computational complexity, memory requirements, and energy consumption, which limit the practical deployment of such models in resource-constrained environments. In response, the research community has increasingly prioritized improving the efficiency and sustainability of deep learning systems, particularly in domains where hardware availability and energy consumption constitute critical constraints.


CNNs have been the dominant paradigm in clinical imaging due to their strong inductive biases, which promote locality and translation invariance. These properties make CNNs particularly data-efficient and robust in scenarios where annotated medical data are scarce. Architectures such as ResNet \cite{he2016deep}, DenseNet \cite{huang2017densely}, MobileNet \cite{howard2017mobilenets}, EfficientNet \cite{tan2019efficientnet} and VGG \cite{simonyan2014very} are widely used as backbones for classification tasks, while U-Net \cite{ronneberger2015u} and its variants have become the standard for medical image segmentation. Among CNN-based approaches, widely adopted architectures include ResNet18, ResNet50, EfficientNet, VGG16, and DenseNet121 \cite{lacci2025deep, kumar2024medical}.

More recently, Transformers have gained increasing attention in clinical imaging for their ability to model long-range dependencies and capture global contextual information. Unlike CNNs, Transformers rely on self-attention mechanisms that enable flexible feature interactions across the entire image, which has proven advantageous for complex medical imaging tasks involving subtle or spatially distributed patterns. Among transformer-based models, Vision Transformer (ViT) \cite{dosovitskiy2020image}, Swin Transformer \cite{liu2021swin}, and Data-efficient Image Transformers (DeiT) \cite{touvron2021training} are the most frequently used architectures in the literature \cite{shobayo2025developments, nauen2025transformer}.

Despite the high diagnostic accuracy achieved by these CNN and Transformer-based architectures, their deployment in clinical settings is frequently restricted due to computational requirements. The significant domain gap between generic pre-training and specialized medical imaging necessitates adaptation strategies that are not only accurate but also environmentally and computationally sustainable. The following subsection details the current landscape of these strategies, ranging from traditional transfer learning to emerging parameter-efficient techniques designed to bridge the domain gap under strict resource constraints.

\subsection{Efficient Training and Optimization Methods}

The increasing computational cost and energy footprint of modern deep learning models have motivated extensive research into training and optimization strategies that improve efficiency without compromising performance \cite{menghani2023efficient}. Rather than focusing exclusively on architectural redesign, many approaches aim to reduce resource requirements through training methodologies that are largely model-agnostic. These methods are particularly relevant in clinical imaging, where limited hardware availability, small datasets, and deployment constraints often preclude full end-to-end training of large models.

\subsubsection{Transfer Learning and Fine-Tuning Strategies}

Transfer learning remains the most widely adopted strategy for efficient training in medical imaging. In this paradigm, models pre-trained on large-scale datasets such as ImageNet are adapted to clinical tasks through fine-tuning. This approach significantly reduces training time and data requirements while improving convergence and generalization. However, full fine-tuning of all model parameters can still be computationally expensive and may lead to overfitting, particularly when medical datasets are small or heterogeneous.

To address these issues, partial fine-tuning strategies are commonly employed, where early layers of the backbone are frozen and only higher-level representations are updated. Layer-wise learning rate decay has also proven effective, especially for Vision Transformers, allowing deeper layers to adapt more strongly to the target domain while preserving general features learned during pretraining. These strategies strike a balance between adaptability and computational efficiency \cite{raghu2019transfusion, gholizade2025review}.

\subsubsection{Parameter-Efficient Fine-Tuning}

More recently, parameter-efficient fine-tuning (PEFT) methods have emerged as a promising alternative to conventional fine-tuning \cite{han2024parameter}. These techniques aim to adapt pre-trained models by introducing a small number of trainable parameters while keeping the majority of the backbone frozen. Representative approaches include low-rank adaptation (LoRA) \cite{hu2022lora}, bias-only tuning (BitFit) \cite{zaken2022bitfit}, and Visual Prompt Tuning (VPT) \cite{jia2022visual}, for transformer architectures.

PEFT methods significantly reduce memory usage, training time, and energy consumption, making them particularly suitable for large Vision Transformer backbones in resource-constrained clinical environments. By limiting the number of trainable parameters, these approaches also reduce the risk of overfitting and improve training stability when only limited labeled data are available.

\subsubsection{Domain Adaptation and Generalization}

Clinical imaging data often exhibit substantial variability due to differences in acquisition devices, imaging protocols, and patient populations. Domain adaptation and domain generalization techniques aim to improve model robustness under such distribution shifts while minimizing the need for extensive retraining.

Common approaches include adversarial learning strategies \cite{wilson2020survey}, self-supervised pretraining on target-domain data \cite{hsu2021robust}, and test-time adaptation \cite{liang2025comprehensive}. While effective, many of these methods require iterative optimization or additional training stages, which may limit their applicability in highly constrained settings. Nonetheless, they play an important role in improving model reliability across institutions and clinical centers.

\subsubsection{Model Compression and Deployment-Oriented Optimization}

Model compression techniques focus on reducing computational complexity after or during training \cite{neill2020overview}. Pruning methods remove redundant weights or structures from a network \cite{he2023structured}, while quantization reduces numerical precision to decrease memory usage and inference latency \cite{gholami2022survey}. Knowledge distillation trains a compact student model to replicate the behavior of a larger teacher model, often achieving competitive performance with significantly reduced model size \cite{gou2021knowledge}.

Although these approaches can yield substantial efficiency gains, they often involve additional training stages or iterative optimization procedures. In clinical contexts, their adoption is typically driven by deployment requirements rather than training efficiency alone.

\section{Proposed methodology}
\label{Sec:methodology}

We propose a resource-efficient training pipeline that aims to maximize classification accuracy while minimizing both computational time and energy consumption. Our framework comprises four complementary stages: 
\begin{enumerate}
    \item Architectural decoupling to separate feature extraction from classification.
    \item Backbone adaptation to align statistical distributions across domains.
    \item Classifier redesign to increase accuracy by enabling more complex decision boundaries.
    \item Targeted training strategies that prioritize informative or ambiguous samples.
\end{enumerate}
Together, these components form a modular and scalable framework suited for resource-constrained environments.

\subsection{Decoupling backbone from classifier}

Classical fine-tune approaches process all data through the whole model during training, extracting features and updating the weights of the classifier and some layers of the backbone, on an end-to-end approach. On most of the CNN architectures, processing data through the backbone is computationally more expensive than doing it through the head \cite{He_2015_CVPR, li2023architecture}. Most fine tuning approaches rely on freezing most of the backbone but some of the last layers, so they require to pass all data on an iterative way in order to update the unfrozen layers. But some other fine tuning approaches, often called feature extraction, freeze the whole backbone and only update the head. This adds a computational overhead, because the backbone is not being updated, and the same features are being extracted on each epoch. 

We propose decoupling the backbone entirely from the classifier, on a two-step procedure. First, all data is processed through the backbone and stored as features. Then, those features would be used to train the classifier independently. This reduces computation time drastically, as the head is significantly lighter than the backbone, and features are only extracted once. 

This approach also allows to extract features on a single step, and reuse them across multiple hyperparameter tuning iterations of the classifier. But this comes with one drawback. The backbone is pretrained on other data, that could have different features than the new domain, and since its weights are not being updated during the iterative training loop, it could lead to a feature domain shift. To mitigate this, we implement a lightweight backbone adaptation phase, detailed below in Section \ref{subsec:backbone_adaptation}.

\subsection{Backbone adaptation}
\label{subsec:backbone_adaptation}

To address the main limitations of model decoupling, domain shift and statistical misalignment, while keeping computational overhead as low as possible, we propose two different approaches, targeting the different normalization layers of CNNs and Transformers.

\subsubsection{CNNs: Thresholded Batch Normalization adaptation}

Since strategies that focus on updating BN layers have shown promising results to adapt models to new data distributions \cite{li2018adaptive}, we introduce an adaptive procedure that updates BN statistics prior to feature extraction. 

The core idea involves forwarding samples through the backbone in training mode while keeping all parameters frozen. By disabling gradient computation, the procedure ensures that only the internal state of the BN layers is updated, while the model weights remain unchanged. This approach enables a fast adaptation step with minimal memory and computational cost.

To avoid unnecessary forward passes, we monitor the evolution of BN statistics during this phase. We extract and concatenate the running mean vectors from all BN layers and compute their normalized difference with respect to the previous iteration, stopping the procedure once the relative change falls under a predefined threshold. This allows to early-stop the adaptation once no more meaningful shifts are detected.

In practice, we implement this by iterating over batches of data, also allowing to define a maximum number of batches as second stopping mechanism. At each step, the normalized change in the running means, denoted $\delta$, is calculated using:

\begin{equation}
\delta = \frac{ \left\| {\mu}^{(t)} - {\mu}^{(t-1)} \right\|_2 }{ \left\| {\mu}^{(t-1)} \right\|_2 + \varepsilon }
\quad \text{with} \quad \delta < \tau_{\text{BN}}
\end{equation}

where $\mu^{(t)}$ represents the concatenated running means at step $t$, and $\varepsilon$ is a small constant added for numerical stability. The adaptation process continues as long as $\delta \geq \tau_{\text{BN}}$, where $\tau_{\text{BN}}$ is a predefined convergence threshold. This strategy offers a practical trade-off between domain adaptation and computational cost by avoiding backpropagation while successfully aligning BN statistics to the target domain.

\subsubsection{Transformers: Thresholded Layer Normalization through bias adaptation}
For Transformers, which typically utilize LN rather than BN, we propose an iterative adaptation strategy that combines the principles of BitFit \cite{zaken2022bitfit} with an early-exit optimization mechanism. Unlike BN, LN layers do not maintain running statistics, therefore, we explicitly align the learnable bias parameters to compensate for covariate shift in the feature space.

For each LN layer, we align the learnable bias parameters to compensate for covariate shift by monitoring the layer's output distribution. Given output activations $x \in \mathbb{R}^{B \times N \times C}$, where $B$ is the batch size, $N$ is the number of patch tokens (sequence length), and $C$ is the embedding dimension, we update the bias parameter $b$:

\begin{equation}
    b^{(t)} = b^{(t-1)} - \eta \cdot \mathbb{E}_{B,N}[x^{(t)}]
\end{equation}

where $\eta$ represents the adaptation rate and $\mathbb{E}_{B,N}[x]$ denotes the mean calculated over the $B \times N$ elements for each feature channel. This update rule effectively centers the distribution of each token's features, compensating for the global shift introduced by the target domain without modifying the attention weights.

While BitFit serves as the foundation for this approach, our implementation introduces three significant modifications focused on efficiency:

\begin{itemize}
    \item \textbf{Gradient-Free Optimization}: Unlike the original BitFit \cite{zaken2022bitfit}, which relies on backpropagation and label-based loss functions, our approach is gradient-free. We perform direct bias shifts based on internal activation statistics to re-center the distribution, eliminating the need for an optimizer or ground-truth labels during adaptation.
    \item  \textbf{Layer-wise Convergence Monitoring}: Instead of training for a fixed duration, we monitor the relative update magnitude for each individual layer. A layer $l$ is considered converged and excluded from subsequent updates when:
    \begin{equation}
        \Delta^{(t)}_l = \| b^{(t)}_l - b^{(t-1)}_l \|_2 < \tau_{\text{LN}}
    \end{equation}
    where $\tau_{\text{LN}}$ is a predefined threshold. This prevents over-fitting to the calibration set and reduces unnecessary updates.
    \item \textbf{Computational Early-Exit}: To minimize GPU cycles, we implement a dynamic hook mechanism. In each iteration, the forward pass is terminated immediately after reaching the deepest non-converged LN layer, avoiding redundant computation in the subsequent frozen layers of the transformer.
    
\end{itemize}

\subsection{Proposed head architecture}

Most CNNs and Transformer architectures rely on a very simple classifier, having only one fully connected (FC) layer. While this design is efficient and effective when relying on updating the backbone weights, it lacks generalization capability when relying only on BN adaptation. 

To address this, we propose a slightly more expressive classification head, composed of the following sequence:

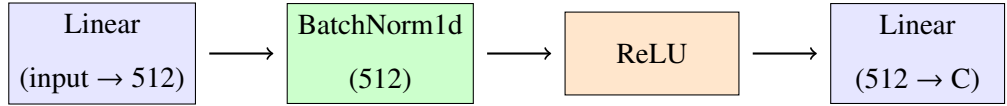
\begin{figure}[h]
\centering
\begin{tikzpicture}[
  node distance=1.2cm,
  every node/.style={draw, minimum height=1.1cm, minimum width=2.3cm, align=center, font=\small},
  relu/.style={draw, fill=orange!20},
  linear/.style={draw, fill=blue!10},
  bn/.style={draw, fill=green!20},
  output/.style={draw, fill=red!20}
]

\node[linear] (fc1) {Linear\\(input → 512)};
\node[bn, right=of fc1] (bn1) {BatchNorm1d\\(512)};
\node[relu, right=of bn1] (relu1) {ReLU};
\node[linear, right=of relu1] (fc2) {Linear\\(512 → C)};

\draw[->, thick, shorten >=5pt, shorten <=5pt] (fc1) -- (bn1);
\draw[->, thick, shorten >=5pt, shorten <=5pt] (bn1) -- (relu1);
\draw[->, thick, shorten >=5pt, shorten <=5pt] (relu1) -- (fc2);

\end{tikzpicture}
\caption{Proposed classifier head, where $C$ denotes the number of output classes.}
\end{figure}

The inclusion of BN and ReLU improves feature conditioning and encourages non-linearity, which can enhance generalization under domain shift. Despite the added complexity, the head remains lightweight and adds negligible overhead. At a standard 224 $\times$ 224 image resolution, the redesigned head tailored for the ResNet50 introduces only 1.05 MFLOPS of overhead. In comparison to the 4.09 GFLOPS required for the complete ResNet50 architecture, this represents just 0.025\% of the total computational complexity.

\subsection{Weighted classifier training}

On the training phase, only the independent classifier head is updated using the features extracted by the backbone. 

To further enhance generalization and focus on learning challenging samples, we adopt a margin-based samples reweighting strategy. The intuition is that not all training samples contribute equally to improving the model, since samples for which the model is already confident are less instructive than those where the decision is ambiguous. Therefore, we aim to emphasize these ``hard" samples by assigning them higher weights during optimization.

For each sample $i$ in a batch of $N$, we determine the prediction margin $m_i$ as the difference between the highest and second-highest predicted class probabilities:

\begin{equation}
m_i = p_i^{(1)} - p_i^{(2)}
\end{equation}

A small margin $m_i$ indicates model uncertainty, suggesting that the sample lies near a decision boundary and is more instructive for the learning process. We identify a threshold hyperparameter $\gamma$, representing a specific percentile of margins within the batch, and assign an amplification factor $\alpha > 1$ to these ambiguous samples. This results in the following weighted cross-entropy loss:

\begin{equation}
\mathcal{L} = - \sum_{i=1}^{N} w_i \cdot \log p_{i,y_i}
\quad \text{where} \quad
w_i = 
\begin{cases}
\alpha, & \text{if } m_i \leq \gamma \\
1, & \text{otherwise}
\end{cases}
\end{equation}

In this formulation, $p_{i,y_i}$ represents the predicted softmax probability for the true class $y_i$. By this logic, samples with a margin below the threshold, representing the most ambiguous predictions, are penalized more heavily via the amplification factor $\alpha$, while all other samples maintain a standard weight of one.



\section{Evaluation}
\label{Sec:evaluation}

In this section, we detail the experimental framework used to evaluate the trade-off between model performance, computational efficiency and environmental sustainability. To validate our model, we employed both accuracy and performance metrics. The total training time was measured, and for our model this includes feature extraction, backbone adaptation and classifier training. Additionally, to measure the environmental footprint of the training sequence, emissions were tracked. To do so, we employed the CodeCarbon\footnote{https://github.com/mlco2/codecarbon} library to measure the CO\textsubscript{2} derived from the conversion of the electrical cost, taking into account the carbon intensity of the local energy grid.


    



\subsection{Datasets}
\label{subsec:datasets}


The choice of the medical domain for evaluating this framework is motivated by its unique combination of high stakes and significant technical constraints. In clinical environments, where access to high-performance hardware is very limited, the adaptation and deployment of computationally intensive methods remains impractical. Consequently, many medical centers are unable to benefit from recent state-of-the-art advancements due to a mismatch between model hardware demands and local resource limitations \cite{lepcha2025deep, patel2024traditional, miotto2018deep}.


To address these challenges, we evaluate the proposed framework across three distinct medical imaging datasets, selected to represent a diverse spectrum of real-world clinical and computational hurdles. This selection benchmarks the framework under four specific conditions:

\begin{enumerate}
\item \textbf{Domain diversity:} Assessing the model's adaptability across contrasting clinical modalities, ranging from macro-scale radiological images to micro-scale histopathological slides.
\item \textbf{Data scarcity:} Evaluating performance in data-constrained regimes with a limited number of samples per class.
\item \textbf{Computational Scalability:} Utilizing a large-scale benchmark to analyze how the trade-off between accuracy and environmental footprint scales with increasing data volume ($N$) and hyperparameter search space ($P$).
\item \textbf{Image complexity and resolution}: Investigating the impact of varying spatial dimensions on memory consumption. By comparing standard resolutions (224 $\times$ 224) against higher resolution images, we evaluate the framework's ability to handle the increased computational demand.
\end{enumerate}


The primary characteristics and the specific role of each dataset within our experimental framework are summarized in Table  \ref{tab:all_datasets}.

\begin{table}[ht]
\centering
\caption{Summary of the medical imaging datasets used for evaluation.}
\label{tab:all_datasets}
\begin{tabular}{@{}lcccc@{}}
\toprule
\textbf{Dataset} & \textbf{Modality} & \textbf{Samples} & \textbf{Classes} & \textbf{Resolution}  \\ \midrule
Brain Cancer MRI & Radiological & 6,056 & 3 & $224 \times 224$ \\
BreakHis & Histopathology & 1,995 & 2 & $672 \times 448$  \\
PatchCamelyon & Histopathology & 327,670 & 2 & $224 \times 224$  \\ \bottomrule
\end{tabular}
\end{table}

\subsubsection{Brain Cancer MRI}

\begin{figure}[ht]
    \centering
    \includegraphics[width=0.18\textwidth]{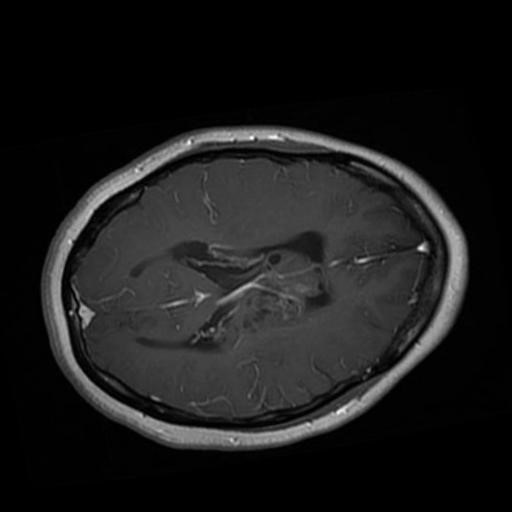}
    \includegraphics[width=0.18\textwidth]{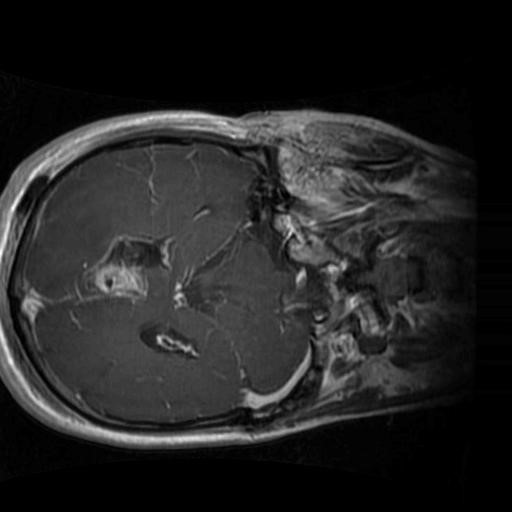}
    \includegraphics[width=0.18\textwidth]{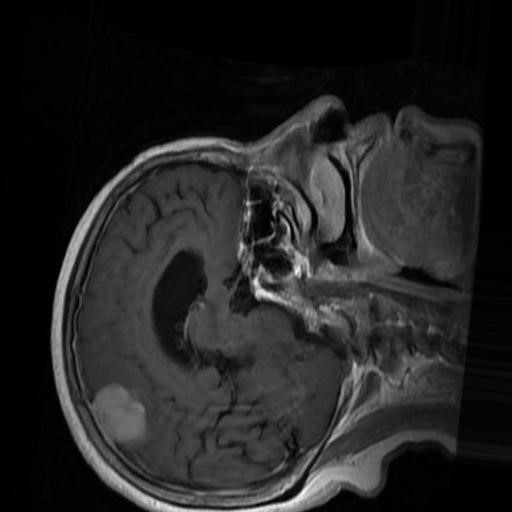}
    \includegraphics[width=0.18\textwidth]{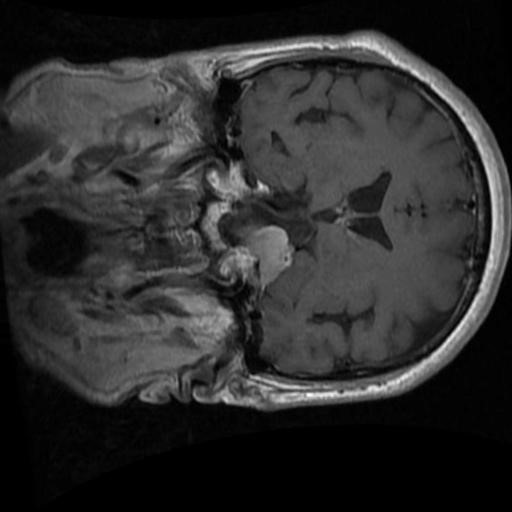}
    \includegraphics[width=0.18\textwidth]{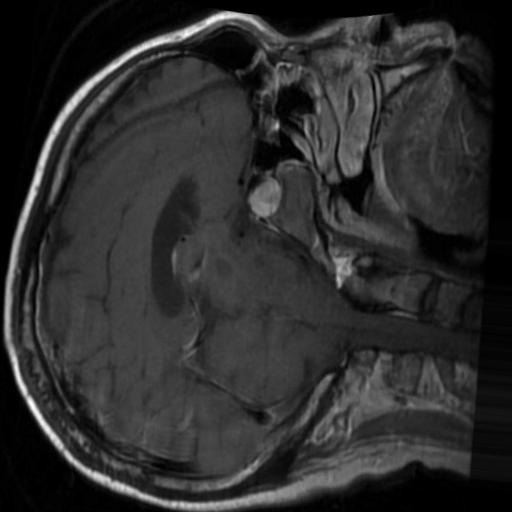}
    \caption{Samples from the Brain Cancer MRI dataset}
    \label{fig:brain_samples}
\end{figure}

The Brain Cancer MRI dataset \cite{rahman2024brain} consist of 6056 grayscale magnetic resonance images (originally \( 512 \times 512 \) and resized to \(224 \times 224\)), used to classify brain tumors into three categories: glioma, meningioma and pituitary tumor. Representative samples from this dataset are presented in Figure \ref{fig:brain_samples}. This dataset serves as a benchmark for evaluating the model's generalizability to non-RGB, radiological modalities.


\subsubsection{BreakHis}

\begin{figure}[ht]
    \centering
    \includegraphics[width=0.18\textwidth]{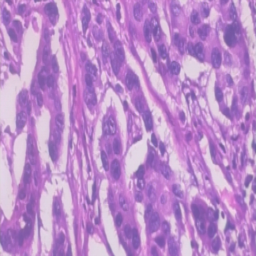}
    \includegraphics[width=0.18\textwidth]{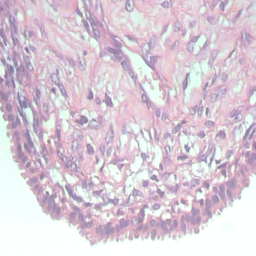}
    \includegraphics[width=0.18\textwidth]{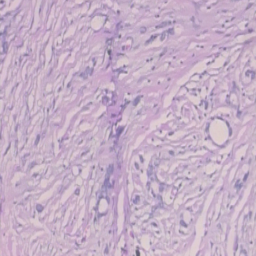}
    \includegraphics[width=0.18\textwidth]{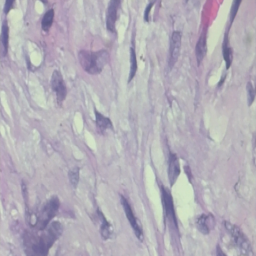}
    \includegraphics[width=0.18\textwidth]{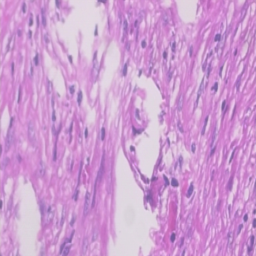}
    \caption{Samples from the BreakHis dataset}
    \label{fig:breakhis_samples}
\end{figure}

The Breast Cancer Histopathological Image Classification (BreakHis) dataset \cite{7312934} contains 1995 microscopic images (originally \(700 \times 460\) and resized to \(672 \times 448\)) of breast tumor tissue collected from 82 patients and divided into malignant and benign. These images are captured at four different magnification factors (\( 40\times\), \(100\times\), \(200\times\) and \(400\times\)). It is characterized by complex cellular patterns, challenging the spatial feature extraction of the models. Representative samples of these histopathological images are shown in Figure \ref{fig:breakhis_samples}. For the purpose of this study, we focus exclusively on the 40$\times$ magnification subset to evaluate model performance on broader architectural patterns of the tissue.


\subsubsection{PCAM}

\begin{figure}[ht]
    \centering
    \includegraphics[width=0.18\textwidth]{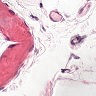}
    \includegraphics[width=0.18\textwidth]{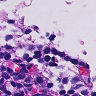}
    \includegraphics[width=0.18\textwidth]{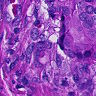}
    \includegraphics[width=0.18\textwidth]{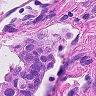}
    \includegraphics[width=0.18\textwidth]{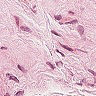}
    \caption{Samples from the PatchCamelyon dataset}
    \label{fig:pcam_samples}
\end{figure}

The PatchCamelyon (PCAM) dataset \cite{veeling2018rotationequivariantcnnsdigital} is a large-scale histopatology benchmark consisting of 327,670 images (originally \(96\times 96 \) and resized to \(224\times 224\)). Representative samples from this dataset are illustrated in Figure \ref{fig:pcam_samples}. Due to its significant volume, PCAM serves as our primary benchmark for the Scalability Analysis, allowing us to observe how models behave as $N$ and $P$ grow.


\subsection{Models}

We evaluate our proposed framework across two dominant architectural para\-digms to assess the efficiency-to-performance ratio of our approach. To provide a rigorous benchmark, we compare our method against the standard adaptation techniques for each architecture type:

\begin{itemize}
    \item \textbf{CNN-based models:} We evaluate ResNet18, ResNet50, MobileNet-V3-Large, and DenseNet121. These models leverage inductive biases like translation invariance and locality. We compare our approach against traditional fine-tuning, where only the last layer of the pre-trained convolutional backbone is updated with the classifier head and the rest of the backbone remains frozen. Full fine-tuning of all backbone parameters was not considered due to its substantial memory and computational requirements on commodity hardware, which conflicts with our target deployment scenario focused on resource-constrained environments.
        
    
    \item \textbf{Transformer-based models:}  We evaluate the Vision Transformer (ViT-B-16), Swin Transformer (Swin-V2-B), and Data-efficient image Transformer (DeiT-Tiny). These models utilize self-attention mechanisms to capture global dependencies. Rather than full fine-tuning, which is computationally exhaustive for large Transformers, we compare our approach against LoRA. By injecting trainable rank-decomposition matrices into the attention layers while keeping the backbone weights frozen, LoRA serves as our state-of-the-art baseline for PEFT.
\end{itemize}

\subsection{Experiments settings}

The experimental framework is designed to evaluate the proposed approach across varying model architectures and operational constraints. CNN experiments were conducted on a NVIDIA RTX 4060 laptop GPU (8GB VRAM) and an AMD Ryzen 7 7435HS CPU. Due to the high memory overhead required for fine-tuning Transformers, these experiments were conducted using an NVIDIA GeForce RTX 3090 Ti GPU (24GB VRAM) and an Intel Core i7-12700K CPU. To ensure reproducibility, we maintain consistent hyperparameters across all runs: ambiguous samples are emphasized using a margin-based weighting strategy with an amplification factor \( \alpha = 10.0 \) and a margin threshold $\gamma$ set to the 20th percentile of the batch. For backbone adaptation, BN statistics are updated using a convergence threshold $\tau_{\text{BN}}$ = 0.01, while the LN bias adaptation is performed with an adaptation rate  $\eta$ = 0.1, a maximum of $t_{\text{max}}$ = 20 steps, and a layer-wise early exit of $\tau_{\text{LN}}$ = 0.0001.

\section{Experimental results}
\label{Sec:results}

In this section, we present a comprehensive analysis of our experimental findings. We evaluated our method based on three primary pillars: predictive performance, computational scalability and hardware efficiency. By benchmarking four CNNs and three Transformers across three datasets we ensure a statistically robust comparison. This section concludes with an ablation study designed to isolate the contributions of the components of the proposed pipeline.

\subsection{Predictive performance}

We first evaluated the predictive accuracy and training efficiency of all seven models across the three selected datasets. This provides a comprehensive performance benchmark against standard fine-tuning and LoRA adaptation strategies. Across all experiments, our method consistently demonstrated a dramatic reduction in computational overhead, often exceeding a 20$\times$ speedup, while maintaining competitive or superior accuracy levels.

On the BreakHis dataset, Table \ref{tab:exp1_breakhis}, our method maintained a high accuracy ($>98\%$) on five out of the seven architectures. Notably, it outperformed traditional fine-tuning on ResNet50 and DenseNet121, as well as LoRA on ViT and Swin, with the latter achieving the highest overall accuracy of 99.35\%. Furthermore, the training efficiency was highlighted by a 21$\times$ speedup on DenseNet121, which reduced CO\textsubscript{2}  emissions by more than 20$\times$. To further validate our method, we followed the protocol described in \cite{saha2024breast}, resizing BreakHis images to 256 $\times$ 256 and averaging the results of independently trained magnification scales. Our pipeline applied to MobileNetV3 achieved an accuracy of 95.59\%, 99.25\%, 99.17\% and 99.04\%, on the corresponding magnifications of $\times$40, $\times$100, $\times$200 and $\times$400, averaging a total accuracy of 99.01\%, surpassing the 98.11\% reported in their study.

\begin{table}[htb!]
\centering
\caption{Performance comparison on the BreakHis dataset. Baseline refers to Fine-Tuning for CNNs and LoRA for Transformers. Best results are highlighted in bold.}
\label{tab:exp1_breakhis}
\small
\begin{tabular}{lcccccc}
\toprule
\multirow{2}{*}{Architecture} & \multicolumn{2}{c}{Accuracy (\%)} & \multicolumn{2}{c}{Time (s)} & \multicolumn{2}{c}{CO\textsubscript{2} (kg)} \\
\cmidrule(lr){2-3} \cmidrule(lr){4-5} \cmidrule(lr){6-7}
 & Baseline & Ours & Baseline & Ours & Baseline & Ours \\
\midrule
\multicolumn{7}{l}{\textit{CNN Architectures}} \\
\addlinespace[2pt]
ResNet18    & \textbf{99.33} & 98.50          & 367.29  & \textbf{42.62}  & 0.0069 & \textbf{0.0008} \\
ResNet50    & 97.66          & \textbf{98.33} & 513.38  & \textbf{59.74}  & 0.0104 & \textbf{0.0012} \\
MobileNetV3 & \textbf{99.00} & 98.67          & 430.98  & \textbf{57.14}  & 0.0087 & \textbf{0.0010} \\
DenseNet121 & 93.46          & \textbf{94.52} & 3970.62 & \textbf{184.37} & 0.0794 & \textbf{0.0038} \\
\midrule
\multicolumn{7}{l}{\textit{Transformer Architectures }} \\
\addlinespace[2pt]
ViT         & 91.71          & \textbf{99.00} & 178.30  & \textbf{33.20}  & 0.0019 & \textbf{0.0004} \\
Swin        & 88.32          & \textbf{99.35} & 196.44  & \textbf{41.92}  & 0.0022     & \textbf{0.0005} \\
DeiT        &      \textbf{95.54}          &    91.43            &    152.41     &  \textbf{19.21}               &   0.00014     &       \textbf{0.0001}          \\
\bottomrule
\end{tabular}
\end{table}

The Brain Cancer MRI results, Table \ref{tab:exp1_braincancer}, further validate our pipeline, where we achieved a peak accuracy of 99.71\% using MobileNetV3. This significantly outperformed the fine-tuned counterpart by nearly 6 percentage points, despite requiring 22$\times$ less training time and 25$\times$ lower CO\textsubscript{2} emissions. Even in high-capacity CNNs like ResNet50 and DenseNet121, our method preserved accuracy within 3\% of the baseline while accelerating training by 5.4$\times$ and 10.7$\times$, respectively.

\begin{table}[htb!]
\centering
\caption{Performance comparison on the Brain Cancer MRI dataset. Baseline refers to Fine-Tuning for CNNs and LoRA for Transformers. Best results are highlighted in bold.}
\label{tab:exp1_braincancer}
\small
\begin{tabular}{lcccccc}
\toprule
\multirow{2}{*}{Architecture} & \multicolumn{2}{c}{Accuracy (\%)} & \multicolumn{2}{c}{Time (s)} & \multicolumn{2}{c}{CO\textsubscript{2} (kg)} \\
\cmidrule(lr){2-3} \cmidrule(lr){4-5} \cmidrule(lr){6-7}
 & Baseline & Ours & Baseline & Ours & Baseline & Ours \\
\midrule
\multicolumn{7}{l}{\textit{CNN Architectures}} \\
\addlinespace[2pt]
ResNet18    & \textbf{99.16} & 98.42          & 196.67  & \textbf{32.48}  & 0.0038 & \textbf{0.0006} \\
ResNet50    & \textbf{99.25} & 97.75 & 311.55  & \textbf{57.41}  & 0.0064 & \textbf{0.0011} \\
MobileNetV3 & 93.93 & \textbf{99.71}          & 770.21  & \textbf{34.78}  & 0.0151 & \textbf{0.0006} \\
DenseNet121 & \textbf{99.58}          & 96.76 & 625.14 & \textbf{58.20} & 0.0127 & \textbf{0.0011} \\
\midrule
\multicolumn{7}{l}{\textit{Transformer Architectures }} \\
\addlinespace[2pt]
ViT         & 96.67          & \textbf{98.24} & 337.72  & \textbf{33.36}  & 0.0055 & \textbf{0.0004} \\
Swin        & 94.92          & \textbf{97.76} & 302.98  & \textbf{42.53}  & 0.0048     & \textbf{0.0006} \\
DeiT        &     \textbf{98.84}           &  93.68              &    193.74     &  \textbf{22.16}               &   0.0018     &   \textbf{0.0001}              \\
\bottomrule
\end{tabular}
\end{table}

Finally, on the PatchCamelyon dataset, Table \ref{tab:exp1_pcam}, our pipeline demonstrated strong robustness. While it performed slightly below standard fine-tuning for CNNs (averaging 1.6\% lower), it significantly surpassed LoRA for Transformers, yielding an average accuracy 4.74 percentage points higher. On the most time demanding models, traditional techniques required about 5 hours to train, while our method completed these trainings in less than half an hour, presenting a reduction of 15.94$\times$ for DenseNet121 and 11.95$\times$ for ViT. This confirms that our approach offers a superior balance of efficiency and predictive power compared to existing adaptation strategies.

\begin{table}[htb!]
\centering
\caption{Performance comparison on the PatchCamelyon dataset. Baseline refers to Fine-Tuning for CNNs and LoRA for Transformers. Best results are highlighted in bold.}
\label{tab:exp1_pcam}
\small
\begin{tabular}{lcccccc}
\toprule
\multirow{2}{*}{Architecture} & \multicolumn{2}{c}{Accuracy (\%)} & \multicolumn{2}{c}{Time (s)} & \multicolumn{2}{c}{CO\textsubscript{2} (kg)} \\
\cmidrule(lr){2-3} \cmidrule(lr){4-5} \cmidrule(lr){6-7}
 & Baseline & Ours & Baseline & Ours & Baseline & Ours \\
\midrule
\multicolumn{7}{l}{\textit{CNN Architectures}} \\
\addlinespace[2pt]
ResNet18    & \textbf{79.78} & 79.03          & 5953.25  & \textbf{841.32}  & 0.1202 & \textbf{0.0163} \\
ResNet50    & \textbf{83.54} & 81.17 & 10191.44  & \textbf{1114.45}  & 0.2140 & \textbf{0.0229} \\
MobileNetV3 & \textbf{82.96} & 82.64          & 8009.09  & \textbf{695.83}  & 0.1665 & \textbf{0.0134} \\
DenseNet121 & \textbf{84.22}          & 80.67 & 20530.34 & \textbf{1287.82} & 0.4382 & \textbf{0.0264} \\
\midrule
\multicolumn{7}{l}{\textit{Transformer Architectures }} \\
\addlinespace[2pt]
ViT         & 84.11          & \textbf{84.29} & 16313.33  & \textbf{1365.29}  & 0.2950 & \textbf{0.0210} \\
Swin        & 81.07          & \textbf{82.22} & 13677.74 & \textbf{1872.48}  & 0.2447     & \textbf{0.0296} \\
DeiT        &      \textbf{86.28}          &    82.26            &    7475.13     &       \textbf{796.30}          &    0.0946    &       \textbf{0.0071}          \\
\bottomrule
\end{tabular}
\end{table}

While our pipeline achieves high accuracy on standard Transformers, the performance degradation observed in DeiT can be attributed to its specialized pre-training. Unlike ViT or Swin, DeiT is a distilled student model that employs a dual-token system, a class token and a distillation token, during its initial training to mimic a CNN teacher. The LN layers in the backbone were optimized to stabilize a feature distribution taking into account both signals. By shifting the bias using raw activation means, we are disrupting the balance between distillation and class tokens.

\subsection{Scalability Analysis}

Second, we evaluated the scalability of our proposed pipeline. A significant advantage of our method lies in its efficiency during hyperparameter optimization or training for a high number of epochs. In traditional approaches, each unique set of hyperparameters requires a complete pass through the entire network architecture across all training epochs. In contrast, our approach only requires to extract features once and then the classifier training is performed independently.

To evaluate this scalability, we conducted a series of experiments using subsets of the PatchCamelyon dataset across varying sizes (1,000, 10,000 and 100,000 samples) and hyperparameter set sizes ($P$). For this analysis, we focused on the CNN architectures. To ensure a controlled comparison, we assumed each hyperparameter set require a fixed 10 epochs of training.

The results, summarized in Table \ref{tab:comprehensive_tuning_time}, show the training time reduction obtained using our model against using standard fine-tuning. As the number of $P$ increases, the percentage of time reduction grows significantly. For the ResNet50 model on a 1$k$ sample dataset, increasing $P$ from 1 to 160 resulted in a time reduction improvement from 83.3\% to a 98.7\%. The most substantial gains are observed in the experiment where DenseNet121 is trained on 100$k$ samples and $P$ = 160, where fine-tuning required 281 hours (more than 11 days), and our approach completed the same task on 1.7 hours. 

To contextualize these benefits using Environmental Protection Agency (EPA) conversion metrics, the 0.105 kg of CO\textsubscript{2} saved during a single training run of our smallest model (ResNet18) is equivalent to avoiding a 0.43 km drive in a standard gasoline-powered vehicle. While this saving may appear modest for a single run, the cumulative impact is substantial when considering the scalability results. For the previously mentioned DenseNet121 experiment, the energy consumed would be equivalent to driving approximately 72 km, while our approach reduces this impact to less than 0.5 km.


\begin{table*}[htb!]
    \centering
    \caption{Comparison of training times (in seconds) across different models using our proposed method and fine-tuning (FT), along with the corresponding percentage reduction in training time (Red.). Results are reported for diverse hyperparameter set sizes ($P$) and different dataset sizes.}
    
    \label{tab:comprehensive_tuning_time}
    \resizebox{\textwidth}{!}{%
    \begin{tabular}{ll|ccc|ccc|ccc}
        \toprule
        \textbf{Model} & \textbf{P} & \multicolumn{3}{c|}{\textbf{1k samples}} & \multicolumn{3}{c|}{\textbf{10k samples}} & \multicolumn{3}{c}{\textbf{100k samples}} \\
        \cmidrule(lr){3-5} \cmidrule(lr){6-8} \cmidrule(lr){9-11}
        & & \textbf{FT} & \textbf{Ours} & \textbf{Red.} & \textbf{FT} & \textbf{Ours} & \textbf{Red.} & \textbf{FT} & \textbf{Ours} & \textbf{Red.} \\
        \midrule
        \multirow{6}{*}{\textbf{Resnet18}}
        & 1   & 21.98 & 3.60 & 83.6\% & 225.26 & 27.41 & 87.8\% & 1865.54 & 253.43 & 86.4\% \\
        & 10  & 216.98 & 7.67 & 96.5\% & 2315.68 & 60.47 & 97.4\% & 18282.29 & 547.36 & 97.0\% \\
        & 20  & 441.36 & 11.81 & 97.3\% & 4375.85 & 99.62 & 97.7\% & 37516.01 & 876.27 & 97.7\% \\
        & 30  & 631.96 & 16.03 & 97.5\% & 6870.67 & 157.24 & 97.7\% & 57831.74 & 1207.41 & 97.9\% \\
        & 40  & 854.48 & 19.94 & 97.7\% & 8945.79 & 232.38 & 97.4\% & 74623.09 & 1588.67 & 97.9\% \\
        & 160 & 3570.07 & 72.67 & 98.0\% & 35727.12 & 574.03 & 98.4\% & 296767.30 & 5762.29 & 98.1\% \\
        \midrule
        \multirow{6}{*}{\textbf{Resnet50}}
        & 1   & 36.53 & 6.10 & 83.3\% & 375.20 & 41.93 & 88.8\% & 3230.34 & 360.77 & 88.8\% \\
        & 10  & 371.20 & 9.76 & 97.4\% & 3712.64 & 75.52 & 98.0\% & 33453.65 & 691.19 & 97.9\% \\
        & 20  & 699.95 & 14.34 & 97.9\% & 7377.38 & 112.42 & 98.5\% & 65823.80 & 1059.39 & 98.4\% \\
        & 30  & 1113.41 & 18.80 & 98.3\% & 11338.70 & 150.02 & 98.7\% & 94613.49 & 1425.01 & 98.5\% \\
        & 40  & 1555.50 & 23.74 & 98.5\% & 14940.20 & 186.21 & 98.8\% & 124909.22 & 1799.00 & 98.6\% \\
        & 160 & 6206.57 & 80.19 & 98.7\% & 58601.66 & 671.48 & 98.9\% & 539911.38 & 6269.61 & 98.8\% \\
        \midrule
        \multirow{6}{*}{\textbf{Mobilenet}}
        & 1   & 28.84 & 3.74 & 87.0\% & 313.78 & 25.51 & 91.9\% & 2994.20 & 360.77 & 88.0\% \\
        & 10  & 294.66 & 6.22 & 97.9\% & 2958.90 & 59.33 & 98.0\% & 29352.11 & 543.28 & 98.2\% \\
        & 20  & 578.24 & 10.16 & 98.2\% & 5985.34 & 97.60 & 98.4\% & 61426.36 & 892.17 & 98.5\% \\
        & 30  & 861.74 & 14.55 & 98.3\% & 9656.08 & 129.43 & 98.7\% & 92575.95 & 1256.01 & 98.6\% \\
        & 40  & 1153.69 & 18.79 & 98.4\% & 12285.33 & 166.00 & 98.7\% & 119785.14 & 1612.47 & 98.7\% \\
        & 160 & 4614.75 & 70.53 & 98.5\% & 48963.08 & 637.71 & 98.7\% & 466367.61 & 5900.43 & 98.7\% \\
        \midrule
        \multirow{6}{*}{\textbf{Densenet121}}
        & 1   & 64.33 & 8.62 & 86.6\% & 691.13 & 45.21 & 93.5\% & 6338.11 & 379.11 & 94.0\% \\
        & 10  & 648.63 & 10.76 & 98.3\% & 6758.94 & 77.44 & 98.9\% & 64445.41 & 700.52 & 98.9\% \\
        & 20  & 1288.74 & 17.41 & 98.6\% & 13365.95 & 113.91 & 99.1\% & 128848.81 & 1059.44 & 99.2\% \\
        & 30  & 1937.32 & 20.45 & 98.9\% & 21194.73 & 149.27 & 99.3\% & 189999.51 & 1414.08 & 99.3\% \\
        & 40  & 2504.92 & 22.99 & 99.1\% & 26270.26 & 183.34 & 99.3\% & 255414.99 & 1772.28 & 99.3\% \\
        & 160 & 9991.81 & 80.90 & 99.2\% & 103584.71 & 622.16 & 99.4\% & 1014054.28 & 6059.69 & 99.4\% \\
        \bottomrule
    \end{tabular}
    } 
\end{table*}

\begin{figure}[htbp]
    \centering
    \includegraphics[width=\textwidth]{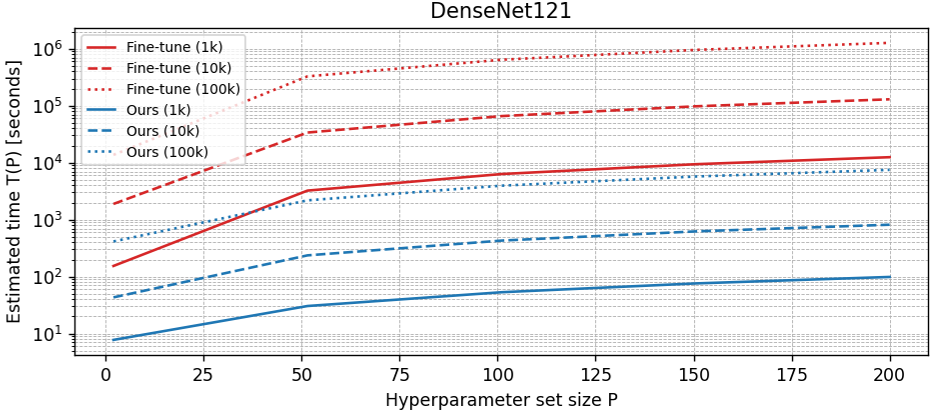}  
    \caption{
        Estimated time required (in seconds) to perform hyper parameter tuning using different amounts of hyper parameter sets and data sizes for Densenet121.
    }
    \label{fig:time_complexity4}
\end{figure}

Table \ref{tab:time_complexity} reports linear regression fits to the empirical training times presented in Table \ref{tab:comprehensive_tuning_time}, where the slope of each fit represents the marginal cost of adding one additional hyperparameter configuration. While both fine-tuning and our proposed approach follow a theoretical complexity of $O(n \cdot P)$, their associated slopes differ substantially. The speedup per configuration, computed as the ratio of fine-tuning slope to ours, reveals that our method scales far more favorably: for DenseNet121 at 100k samples, fine-tuning requires an additional 1.7 hours per new configuration, while our method adds only 35 seconds, a speedup of 177$\times$. Notably, this speedup grows with the fine-tuning cost of the backbone rather than raw model complexity. DenseNet121, whose dense connectivity pattern makes iterative fine-tuning disproportionately expensive, consistently shows the largest speedup across all dataset sizes. DenseNet121 approximations are visually represented in Figure \ref{fig:time_complexity4}.

\begin{table}[ht]
\centering
\renewcommand{\arraystretch}{1.3}
\caption{Linear regression fits to empirical training times from Table \ref{tab:comprehensive_tuning_time}, reporting estimated time per hyperparameter configuration (slope) and marginal speedup per additional configuration across models and dataset sizes.}
\resizebox{\textwidth}{!}{
\begin{tabular}{cccccc}
\hline
\textbf{Model} & \textbf{Approach} & \textbf{T(P) for 1k} & \textbf{T(P) for 10k} & \textbf{T(P) for 100k} \\
\hline
ResNet18    & Fine-tune & $22.369 \cdot P - 16.930$ & $223.114 \cdot P + 37.923$  & $1852.913 \cdot P + 545.957$ \\
            & Ours      & $0.434 \cdot P + 3.057$   & $3.386 \cdot P + 44.551$    & $34.773 \cdot P + 193.281$ \\
            \cdashline{2-5}
            & Speedup      & 52$\times$    & 66$\times$    & 53$\times$ \\
\hline
ResNet50    & Fine-tune & $38.971 \cdot P - 31.382$ & $365.737 \cdot P + 148.087$ & $3384.860 \cdot P - 3584.411$ \\
            & Ours      & $0.469 \cdot P + 5.105$   & $3.977 \cdot P + 33.256$    & $37.194 \cdot P + 316.051$ \\
            \cdashline{2-5}
            & Speedup      & 83$\times$    & 92$\times$    & 91$\times$ \\
            
\hline
MobileNet   & Fine-tune & $28.831 \cdot P + 1.175$  & $306.000 \cdot P + 49.434$  & $2905.275 \cdot P + 2370.756$ \\
            & Ours      & $0.426 \cdot P + 2.131$   & $3.861 \cdot P + 17.959$    & $35.729 \cdot P + 183.325$ \\
            \cdashline{2-5}
            & Speedup      & 68$\times$    & 79$\times$    & 81$\times$ \\
\hline
DenseNet121 & Fine-tune & $62.292 \cdot P + 29.602$ & $644.711 \cdot P + 599.374$ & $6332.636 \cdot P + 1047.185$ \\
            & Ours      & $0.460 \cdot P + 6.849$   & $3.902 \cdot P + 35.504$    & $35.726 \cdot P + 343.438$ \\
            \cdashline{2-5}
            & Speedup      & 135$\times$    & 165$\times$    & 177$\times$ \\
\hline
\end{tabular}
}
\label{tab:time_complexity}
\end{table}

\subsection{Hardware efficiency}

Third, we evaluated the hardware efficiency of our method to determine if it could remain practicable in resource-constrained environments lacking dedicated GPU acceleration. In traditional approaches, the requirement for backpropagation through millions of parameters makes CPU training prohibitively slow for all but the smallest models.

\begin{figure}[htbp!]
    \centering
    \includegraphics[width=\textwidth]{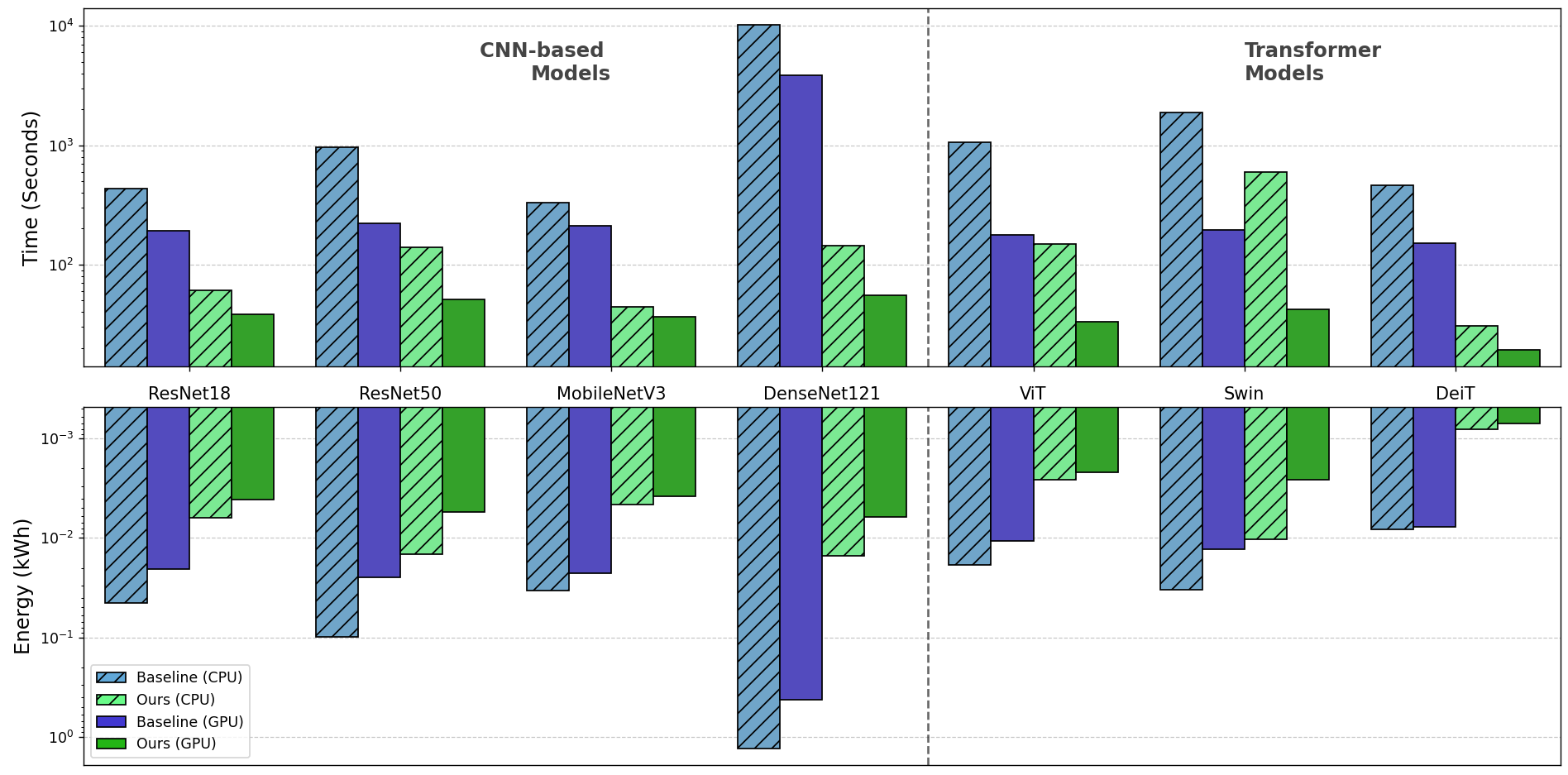}  
    \caption{
        Training comparison using CPU and GPU.
    }
    \label{fig:cpu_gpu}
\end{figure}

Figure \ref{fig:cpu_gpu} illustrates the differences between traditional baselines and our method across CPU and GPU environments. For all models but Swin, our approach executed on a CPU consistently outperformed the baselines running on a GPU. By enabling CPU-based training to surpass the speeds of traditional GPU-accelerated baselines, we demonstrate that training with the proposed approach is practicable even in hardware-constrained environments.

The Swin model presents higher training times on CPU when using our adaptation method. This is primarily because Swin’s hierarchical architecture contains roughly twice the number of LN layers compared to the standard ViT. On a CPU, where operations are largely sequential, each additional layer adds a linear overhead to the adaptation loop. Using a GPU, however, this performance gap is reduced, thanks to the GPU's parallelism to process the increased layer count and Swin's window-shifting operations.

\subsection{Ablation study}

Lastly, we performed an ablation study to determine the individual contribution of each component within our adaptation framework, summarized in Table \ref{tab:ablation_compact}. The base configuration consists of a frozen backbone with the redesigned head trained using standard cross-entropy loss (amplification factor $\alpha$ = 1 and no backbone adaptation). From this base, we isolate two components: backbone adaptation, evaluated without weighted loss (w/o Weight), and weighted loss ($\alpha$ = 10), evaluated without backbone adaptation (w/o Adapt). The Full configuration combines both components. 

For all models, the full pipeline outperforms either component in isolation, confirming that both backbone adaptation and weighted training contribute independently to accuracy improvement. Regarding computational cost, weighted loss introduces negligible overhead. Backbone adaptation is the dominant source of overhead, particularly for Transformer architectures where the iterative LN bias adjustment across many layers accumulates a significant cost, most notably for Swin, which contains roughly twice the number of LN layers compared to ViT. Nevertheless, even in the worst case, the total overhead of the full pipeline remains orders of magnitude lower than traditional fine-tuning, preserving the efficiency advantage that motivates the entire framework.

\begin{table}[htb!]
\centering
\caption{Ablation study of the proposed pipeline components. Performance gains are measured relative to a base configuration (no adaptation and standard training). The \textit{w/o Adapt.} and \textit{w/o Weight} columns isolate the impact of each respective component, while \textit{Full} represents the complete proposed pipeline. Best accuracy improvement values are highlighted in bold.}
\label{tab:ablation_compact}
\small
\begin{tabular}{lcccccc}
\toprule
\multirow{2}{*}{Architecture} & \multicolumn{3}{c}{Accuracy Improvement (\%)} & \multicolumn{3}{c}{Time Overhead (s)} \\
\cmidrule(lr){2-4} \cmidrule(lr){5-7}
 & w/o Adapt. & w/o Weight & Full & w/o Adapt. & w/o Weight & Full \\
\midrule
\multicolumn{7}{l}{\textit{CNN Architectures}} \\
ResNet18    & 1.33 & 1.65 & \textbf{2.33} & 0.14  & 8.54  & 11.99 \\
ResNet50    & 0.45 & 0.78 & \textbf{2.11} & 0.14  & 14.81 & 15.88 \\
MobileNetV3 & 0.66 & 0.21 & \textbf{1.33} & 0.08  & 9.29  & 5.01  \\
DenseNet121 & 1.25 & 2.00 & \textbf{2.43} & 0.00  & 7.55  & 17.16 \\
\midrule
\multicolumn{7}{l}{\textit{Transformer Architectures}} \\
ViT         & 1.00 & 1.33 & \textbf{2.33} & 33.48 & 169.88& 158.73\\
Swin        & 0.30 & 0.64 & \textbf{1.22} & 16.49 & 352.01& 336.55\\
DeiT        & 0.93 & 1.09 & \textbf{1.13} & 0.61  & 12.02 & 11.99 \\
\bottomrule
\end{tabular}
\end{table}

\section{Conclusions}
\label{Sec:conclusions}

In this work we presented a pipeline designed to perform transfer learning on domains where time and energy resources are limited, such as fast prototyping or hardware constrained environments. By decoupling feature extraction and performing lightweight adaptation of the backbone with minimal overhead, our approach eliminates the need for expensive end-to-end backpropagation. These adaptations accompanied with a redesigned, more expressive classification head and a margin-based weighted loss, results in a pipeline that maintains high predictive performance with minimal overhead. This indicates that adapting normalization layers is a highly effective way to bridge the domain gap between generic pre-training and specialized medical domains without the requirement of backpropagation through the backbone.

The experimental results demonstrate that the proposed method offers significant advantages over traditional fine-tuning. We achieved dramatic reductions in training time and CO\textsubscript{2} emissions, achieving up to a 20$\times$ speedup in some cases, while maintaining competitive accuracy across diverse architectures and medical datasets. Specifically, for Transformer architectures, our method consistently outperformed LoRA in terms of both efficiency and accuracy, with the sole exception of the DeiT-Tiny model. The lower performance in that specific case is attributed to DeiT’s specialized distilled pre-training, where LN layers are tuned to balance class and distillation tokens, a balance that our lightweight adaptation mechanism disrupts.

Ultimately, this study highlights a sustainable trade-off: while standard fine-tuning may yield marginal accuracy gains in certain scenarios, the substantial improvements in efficiency and environmental sustainability make our approach a superior choice for large-scale hyperparameter optimization and deployment in resource-scarce settings. A key finding of this research is that our pipeline enables the training of heavy models without the requirement of dedicated GPUs, as our proposal achieves faster training times on a CPU than fine-tuning and LoRA using a GPU, for all but the Swin model. This framework provides a practical pathway for integrating state-of-the-art deep learning into real-world medical diagnostics without relying on high-end industrial computing.

Several avenues for future research emerge from this study. First, while the current work focuses on image classification, extending the decoupled pipeline to other vision tasks such as object detection and segmentation would broaden its applicability. Second, the decoupled nature of our pipeline makes it particularly well-suited for federated learning settings, where data cannot leave the local institution due to privacy regulations. Because the backbone remains frozen and features are precomputed independently, collaborative model training can be achieved by sharing only optimized head parameters or feature distributions rather than raw datasets. This significantly reduces communication overhead and eliminates the need for transferring sensitive information across networks. Third, the current margin-based weighting strategy relies on fixed hyperparameters such as the amplification factor and the percentile threshold, which may not generalize optimally across all scenarios. Replacing these with a dynamic weighting scheme that adapts to the observed margin distribution during training would reduce the need for manual tuning and improve robustness across diverse datasets and architectures.

\section*{Acknowledgements} 
This work was supported by the Ministry of Science and Innovation of Spain (Grant PID2023-147404OB-I00 / AEI / 10.13039 / 501100011033) and together with ``NextGenerationEU''/PRTR by the Ministry for Digital Transformation and Civil Service under grant TSI-100925-2023-1 and by Xunta de Galicia (Grants ED431G 2023/01 and ED431C 2022/44).

\bibliographystyle{elsarticle-num} 
\bibliography{cas-refs}

\end{document}